\documentclass[10pt,twocolumn,letterpaper]{article}

\usepackage[pagenumbers]{cvpr} 

\usepackage{graphicx}
\usepackage{amsmath}
\usepackage{amssymb}
\usepackage{booktabs}
\usepackage{multirow}
\usepackage{array}
\usepackage{pbox}

\usepackage{algorithm2e}
\SetKwComment{Comment}{/* }{ */}
\RestyleAlgo{ruled}

\usepackage[pagebackref,breaklinks,colorlinks]{hyperref}

\usepackage[capitalize]{cleveref}
\crefname{section}{Sec.}{Secs.}
\Crefname{section}{Section}{Sections}
\Crefname{table}{Table}{Tables}
\crefname{table}{Tab.}{Tabs.}

\newcommand{\photo}[1]{%
    \includegraphics[width=3.5cm]{#1}
}
\newcommand{\phototwo}[1]{%
    \includegraphics[width=2cm]{#1}
}

\def\cvprPaperID{11389} 
\def\confName{CVPR}
\def\confYear{2022}

\title{$\mu$NCA: Texture Generation with Ultra-Compact Neural Cellular Automata}

\author{Alexander Mordvintsev\footnotemark \hspace{10mm} Eyvind Niklasson\footnotemark[\value{footnote}] \\
Google Research\\
{\tt\small \{moralex, eyvind\}@google.com}
}

\begin{document}
\twocolumn[{%
\renewcommand\twocolumn[1][]{#1}%
\maketitle
\begin{center}
\includegraphics[width=0.95\linewidth]{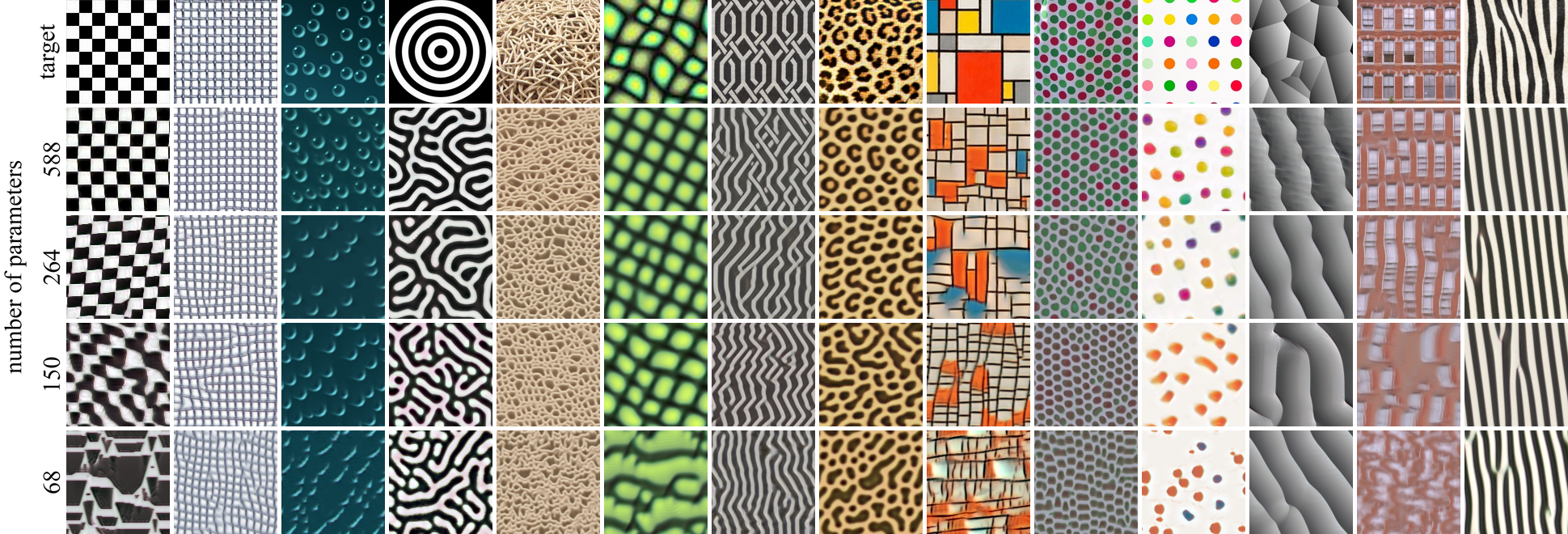}
\end{center}
}]

\maketitle

\renewcommand*{\thefootnote}{\fnsymbol{footnote}}
\footnotetext{* Contributed equally.}
\renewcommand*{\thefootnote}{\arabic{footnote}}

\begin{abstract}
We study the problem of example-based procedural texture synthesis using highly compact models. Given a sample image, we use differentiable programming to train a generative process, parameterised by a recurrent Neural Cellular Automata (NCA) rule. Contrary to the common belief that neural networks should be significantly over-parameterised, we demonstrate that our model architecture and training procedure allows for representing complex texture patterns using just a few hundred learned parameters, making their expressivity comparable to hand-engineered procedural texture generating programs. The smallest models from the proposed $\mu$NCA family scale down to 68 parameters. When using quantisation to one byte per parameter, proposed models can be shrunk to a size range between 588 and 68 bytes. Implementation of a texture generator that uses these parameters to produce images is possible with just a few lines of GLSL\footnote{Interactive WebGL implementations of our NCA rules are available on ShaderToy: \href{https://www.shadertoy.com/view/slGGzD}{NCA-68}, \href{https://www.shadertoy.com/view/styGzD}{NCA-588}} or C code. 
\end{abstract}

\section{Introduction}
\label{sec:intro}

Finding the shortest program to implement a desired behaviour has been of interest both in academic literature and in general computing for almost as long as general purpose computing has been available to the public. Algorithmic information theory defines Kolmogorov complexity \cite{Wikipedia_contributors2021-xb,Kolmogorov1998-wo} as the shortest computer program generating a certain output. There exist parallels to the neighbouring fields of compression and information theory, which handle the more specific instance of transmitting information using the fewest number of bits. Outside academia, work in the demo-scene \cite{Wikipedia_contributors2021-jm} has established a long tradition of encoding complex geometries and rendering algorithms into extremely short programmatic code and small compiled binaries. Likewise, the practice of code-golf - implementing a specified algorithm in the shortest length of source code - is a well recognized form of competition, with worldwide tournaments often involving very creative uses of language constructs and compiler features \cite{Wikipedia_contributors2021-ka}. However, the interest in expressing an algorithm in the shortest form possible is at odds with deep learning approaches du jour, where the trend is towards ever larger models and over-parameterisation \cite{Brown2020-zx,Nakkiran2019-cx}.  

Texture synthesis is usually described as algorithmic generation of new images that express perceptual similarity to a provided texture sample. The most successful non-parametric methods rely on copying target image patches or pixels into the image being generated \cite{Efros_undated-zi,Lefebvre2006-mw}. In recent years, methods relying on gradient-based optimization of image pixels to match target image statistics in a pre-trained neural network have become common \cite{Gatys2015-yl,Heitz2021-bz}. These methods have been extended to optimising the parameters of a feed-forward neural network acting as a generator, transforming a noise vector into images similar to the target texture \cite{Ulyanov2017-af,Li2017-na}. Such feed-forward models can be seen as learned parametric procedural texture generators, typically having $10^5-10^6$ parameters. Such a large parameter space reflects the observation that neural network training achieves best results in the overparameterised regime \cite{Belkin2019-nr,Nakkiran2019-cx}.

In parallel, the computer graphics community has made great progress in building tools for procedural texture generation, allowing artists and designers to creatively combine various primitives into pipelines producing high fidelity textures and materials. Procedural materials have a number of attractive properties, such as efficient sampling and compactness of the underlying representations. The expressiveness of algorithmic texture generators is hard to match: sometimes it takes an experienced programmer just a few hundred characters of OpenGL Shading Language (GLSL) code to create a complex world on a computer screen \cite{Ebert1995-jj,Dong2020-jk}

\section{Related Work}

\subsection{Texture Synthesis}

There exist a few works that attempt to automatically design compact image generators given an image sample. A notable example is work \cite{Reynolds2011-sk} by Craig Reynolds, who used genetic algorithms (based on the work by Karl Sims \cite{Sims1991-re}) driven by human evaluators to produce texture patches that would be unnoticeable on a given background. Some success was also achieved with describing a texture by a relatively small number of statistical parameters (\eg \cite{Portilla2000-ok} uses 710), and synthesising images that match these parameters. Later these ideas were combined with neural network based features to differentiably tune parameters of predefined procedural texture generators to match the provided texture samples \cite{Shi2020:ToG}.

\subsection{Neural Cellular Automata}

Cellular Automata (CA) have been long known to possess the capacity to act as powerful texture generators \cite{Turk1991-nv,Witkin1991-tp}. There is even evidence for cellular automata-like processes generating the skin textures of certain animals \cite{Manukyan2017-zn}. 

Recent work \cite{Niklasson2021-qu} demonstrated a method to learn texture-generating CA given individual image samples. In this method, the CA update rule is parameterised with a comparatively small neural network. The output of the CA is fed into a pre-trained discriminative network. The rules are then differentiably optimised so as to induce feature statistics in certain layers in the discriminative network matching the target statistics of the desired sample image. This model was termed Texture Neural Cellular Automata (TNCA). 

In this work we extend the TNCA line of research through reduction - we study the limits of expressivity of TNCA as they are made more compact. We demonstrate that TNCA having just a few hundred learned parameters (our smallest model has just 68), instead of several thousands, can be trained to generate a wide range of textures. The parameters of the NCA update rule can be seen as a very compact texture representation, and training such models is a form of texture compression. 

\subsection{Optimal Transport}

Optimal transport describes the task of transforming one distribution to another, while minimising the so-called "earth movers distance".  Optimal transport has its roots in the problem setting of finding the optimal redistribution of troops from their current locations to front-line locations, but has since been generalised to transporting a given distribution of points in any space to another distribution under an arbitrary transport cost-matrix. 

Traditionally the problem has been computationally complex, $O(N^4)$ in the number of points. However, recent work \cite{Cuturi2013-sk} by Cuturi et. al have derived a more efficient, and differentiable, algorithm based on the Sinkhorn algorithm, making feasible a whole new set of applications for optimal transport theory \cite{Peyre2018-pl}. The problem setting of computing divergence between two distributions, and differentiating through such a divergence appears in many sub-fields of machine learning \cite{Onken2020-hn}\cite{Torres2021-yd}.

\section{Training Ultra-Compact Neural CAs}

\subsection{Ultra-Compact Neural CA models ($\mu$NCA)}

Our texture generator model is inspired by the Neural CA model from \cite{Niklasson2021-qu,Mordvintsev2021-pw}. It operates on a regular grid where each cell is represented using a vector of scalar values, the first 3 of which are interpreted as the R, G and B color channels, respectively. Similarly, we use differentiable optimization to find a CA rule whose repeated application in each cell on the grid produces the desired texture. The inputs to the update rule on a given cell are only the state of the cell itself and information about its immediate neighbourhood, obtained by convolving a set of fixed filters over the states of surrounding cells. At first glance, this approach may seem contrived or artificially constrained, but it enables surprising and noteworthy learned procedural behaviour. The models used in the works \cite{Mordvintsev2021-pw,Niklasson2021-qu} are already relatively small in size as compared to contemporary work in deep learning. Comprised of just 5856 parameters they are capable of producing a large variety of textures. Authors also demonstrate these models to be tolerant to 8-bit quantization of weights and activations.

We propose a modified version of the texture NCA model from \cite{Niklasson2021-qu}, which we term $\mu$NCA. This model can be summarized in a following set of equations:
\begin{align} 
&\mathbf{s} = [s^0=R, s^1=G, s^2=B, s^3, ... , s^{C-1}]  \label{eq:s} \\
&K =  [\underbrace{K_{lap} \cdots}_{\times N_{lap}}, \underbrace{K_{x} \cdots}_{\times N_{x}},  \underbrace{K_{y} \cdots}_{\times N_{y}}] \label{eq:K} \\
&\mathbf{p} =  \mathit{concat}(\mathbf{s}, [K_i \ast s^i]_{i \in 0 .. C-1}) \label{eq:p} \\
&\mathbf{y} = \mathit{concat}(\mathbf{p}, \mathit{abs}(\mathbf{p})) \label{eq:y} \\
&\mathbf{s}_{next} = \mathbf{s}+\mathbf{y}\mathbf{W}_{C \cdot 4 \times C}+\mathbf{b} \label{eq:snext}
\end{align}

Like TNCA, our $\mu$NCA operates on a uniform grid, where the state of each cell is represented by a $C$-dimentional vector $\mathbf{s}$, and its first three components are visible as RGB colors (\cref{eq:s}), $\mathbf{p}$ represents the perception vector and $\mathbf{s}_{next}$ is the updated cell state after one iteration. The following modifications were made to compress the model, making its memory footprint smaller and evaluation more computationally efficient:

\begin{itemize}
\item The original texture NCA computed a "perception vector" $\mathbf{p}$ by applying a set of convolutional filters ($K_{lap}, K_x, K_y$) to each channel of the cell state. We use only a single filter per channel (repeating filters to cover all channels, see table \ref{tab:config}), thus reducing the dimensionality of $\mathbf{p}$ to be the same as the number of channels. We also concatenate the filtered values with $\mathbf{s}$ so that cells continue to have full access to their own state (\cref{eq:K,eq:p}).

\begin{center}
\scalebox{.75}{\begin{math}\begin{array}{ c c c }
\begin{bmatrix}
1 & 2 & 1\\2 & -12 & 2 \\1 & 2 & 1 \\
\end{bmatrix}
&
\begin{bmatrix}
-1 & 0 & 1\\-2 & 0 & 2 \\-1 & 0 & 1 \\
\end{bmatrix}
&
\begin{bmatrix}
-1 & -2 & -1\\ 0 & 0 & 0 \\1 & 2 & 1 \\
\end{bmatrix}
\\
K_{lap} & K_x & K_y
\end{array}\end{math}}
\end{center}

\item We introduce non-linearity into the model by expanding the perception vector, concatenating it with a vector consisting of the corresponding absolute values of $\mathbf{p}$ (\cref{eq:y}). This can be seen as variant of PReLU non-linearity \cite{He2015-tk}, where both slopes for negative and positive input values can be learned \footnote{for instance, consider ReLU, which can be represented using the linear combination: $\mathit{relu}(x) = (x+|x|)/2$}.

\item We remove the latent 1x1 convolution layer from the model. The cell state updates are instead directly computed with a single matrix multiplication (in practice implemented as 1x1 convolution). All learned parameters are represented by the weight matrix $\mathbf{W}$ and the bias vector $\mathbf{b}$. Total number of parameters equals to $4\cdot C^2+C$, where $C = N_{lap}+N_x+N_y$. Varying $N_{\{lap,x,y\}}$ we define a family NCA of models of different expressive power. Please refer to the \cref{tab:config} for the specific values used in our experiments.

\item Texture NCAs as described in \cite{Niklasson2021-qu} use stochastic updates to introduce randomness into the system as well as to break symmetry between cells. In this work we removed stochastic updates from the CA rule for simplicity. In order to break the symmetry, and to introduce stochasticity in the procedural generation, we initialize the grid with uniformly distributed random values in $[-0.5, 0.5]$.

\end{itemize}

\subsection{Optimal Transport as a Differentiable Texture Similarity Function}

In order to use differentiable optimisation to synthesize textures or to train texture generators, one must define an objective function that would measure the perceptual proximity of the current solution to the target example. The naive pixel-wise difference is not appropriate because by the nature of the problem we are looking to produce novel samples of the target texture, not pixel-perfect copies. Over time, a number of differentiable texture similarity functions have been proposed in literature.

Most contemporary texture generation methods rely on pre-trained convolutional neural networks to extract target image features \cite{Gatys2015-yl} (sometimes referred to as "statistics"), and try to match the distribution of these features between the image under generation and the target texture. A prominent example is the work \cite{Gatys2015-yl}, which used an ImageNet-pretrained VGG feature extractor \cite{Simonyan2014-er} and the difference between the feature channel covariance matrices as the loss. Subsequent works have explored other approaches to computing the discrepancy between feature distributions, more recently based on optimal transportation theory \cite{Houdard2020-av,Heitz2021-bz}.

There are significant drawbacks to VGG-based feature extractors that constrain their applicability. Models using them as feature extractors are unlikely to work well on out-of-distribution data (with regards to the distribution of data the feature extractor was trained on). Similarly, models are primarily constrained to RGB images, preventing the synthesis of textures with additional data encoded in the other channels, such as surface normals, roughness, or opacity. Pre-deep learning methods, such as using texture image patches partially offset many of these issues and may be used in the place of features extracted from deep neural networks \cite{Webster2018-eg,Houdard2020-av}.

In this work we propose a differentiable texture similarity function we term OTT-Loss, combining ideas from \cite{Webster2018-eg,Houdard2020-av}. Our proposed method represents a target texture as a collection of image patches extracted from the image at various scales. A procedure that allows matching the distribution of patches between a source and a target in a differentiable fashion allows for any differentiable model to be used a generator. In practice, we use recent advances in optimal transport theory - specifically the sinkhorn iteration algorithm \cite{Cuturi2013-sk} to calculate the cost of optimal re-allocation of patches between the output of the $\mu$NCA and the distribution of patches from the chosen target texture. 

In this work we extract all possible $K\times K$ patches from the levels of the Gaussian pyramid, computed for the target image. We observed that local contrast enhancement is beneficial for the $\mu$NCA training convergence speed. To achieve it we do the sharpening of the pyramid levels before extracting the patches using the formula:
$$sharpen(I)=I+2(I-G_{5\times 5}\ast I)$$
where $G_{5\times 5}$ is a Gaussian blur kernel.

We flatten each patch into a one-dimensional vector, effectively placing it in a feature space of all possible $K$ sized patches. Our feature extraction procedure is shown in \cref{alg:features}. We then match the induced distributions of patches in this space to patches sampled from the chosen target texture. This must be done separately at each pyramid level:

$$textureLoss(F, F_{target})=\sum_{level} OT_\epsilon(S(F^{level}), S(F_{target}^{level})) $$

where $OT_\epsilon$ denotes the optimal transport divergence, and the function $S(F)$ stochastically subsamples a number of rows (we use 2048) from a given matrix to make the OT computation tractable.

\begin{algorithm}
\caption{$extractFeatures$ function}\label{alg:features}
\KwData{Image $I$, $N_{levels}$, patch size $K$}
\KwResult{$[F^l]_{l\in1..N_{levels}}$}
$I_1 \gets I$\;
\For{$l=1$ \KwTo $N_{levels}$} {
  $I_{sharp} \gets sharpen(I_{l})$ \;
  $F^l \gets$ list of all $K\times K$ overlapping patches extracted from $I_{sharp}$ \;
  $I_{l+1} \gets$ downscale $I_l$ by a factor of 2
}
\end{algorithm}

\subsection{Training procedure}

Our training procedure (\cref{alg:train}) is based on the one proposed in Mordvintsev et. al with a few modifications:

\begin{itemize}
    \item On back-propagation we apply gradient normalization after every CA step. This may be seen as a variant gradient clipping \cite{Pascanu2012-us}, a standard practice to prevent exploding or vanishing gradients in RNN training.

    \item We inject seed states into training batches every $seed\_rate$ training steps, rather than at every step. This allows CA to witness longer CA evolution sequences during training, which results in better long-term pattern quality and stability. The expected number of CA steps seen during training becomes $N_{batch}*N_{steps}*seed\_rate$.
    
    \item Overflow loss, $L_{overflow}$ (see \ref{alg:train}), to keep latent channels in $[-1, 1]$. This stabilises training by preventing drift in latent channels and aids in post-training quantisation. 
\end{itemize}

We keep other augmentations as introduced in \cite{Niklasson2021-qu}, including gradient normalisation prior to taking the optimization step, and using a sample pool to achieve longer-time stabilisation. 

\begin{algorithm}
\caption{Texture NCA Training}\label{alg:train}
\KwData{Target texture image $I_{target}$}
\KwResult{final CA parameters $\theta$}
$\theta \gets$ init CA parameters \;
$opt \gets$ init Adam optimizer\;
$pool \gets$ init state pool with $N_{pool}$ seed states\;
$F_{target} \gets extractFeatures(I_{target})$ \;
\For{$step=1$ \KwTo $N_{train}$}{
  $x \gets$ sample $N_{batch}$ states from $pool$ \;
  \If{$step \bmod seed\_rate=0$}{
    $x[0] \gets $ random seed state
  }
  \For{$i=1$ \KwTo $N_{steps}$}{
    $x = ca(x)$ \;
    $x = norm\_grad(x)$
  }
  $I_{gen} \gets$ extract RGB-channels from $x$\;
  $F_{gen} \gets extractFeatures(I_{gen})$ \;
  $L_{texture} \gets textureLoss(F_{gen}, F_{target})$ \;
  $L_{overflow} \gets \sum |x-clip_{[-1,1]}(x)|$ \;
  $L = L_{texture} + L_{overflow}$ \;
  $\nabla L_\theta \gets$ use back-propagation through CA steps to computes the gradient of $L$\;
  $G_{normed} \gets \nabla L_\theta / ||\nabla L_\theta||_2 $ \;
  $\theta \gets opt.update(G_{normed})$ \;
  place final grid states $x$ back to the $pool$\;
}
\end{algorithm}

\section{Experiments} \label{sec:exp}

\subsection{Datasets}
We demonstrate qualitative performance on an assortment of textures sourced from several datasets. Most of the datasets were released in tandem with texture-generation methods, and have over time been re-used and mixed across different publications, making proper attribution difficult. The textures we use in this work are curated from the following datasets:
\begin{itemize}
  \item Images collected by Portilla and Simoncelli for their work on texture synthesis \cite{Portilla2000-ok}. In turn, these include 'personal photographs' scanned by the authors, as well as the MIT Vision Texture database \cite{noauthor_undated-hk} and scanned images from Textures: A Photographic Album for Artists and Designers \cite{Brodatz1981-ky}.  
  \item Images release together with \emph{Appearance-space texture synthesis} \cite{Lefebvre2006-kh}.
  \item The Describable Textures Dataset\cite{cimpoi14describing}.
\end{itemize}

\subsection{Results}

The opening figure shows curated samples of the $\mu$NCA trained on various textures, some natural and some artificial, for different model sizes of the compact NCA. The number of parameters and filter configuration for each model size is listed in \ref{tab:config}. Notice that the largest model is 588 parameters, or about 2,352 bytes in size, while the smallest model is just 272 bytes. Thorough experimentation with quantisation of the models to just one byte per parameter showed no evidence of degradation in quality. We release a reference shader-based implementation of $\mu$NCA inference which operates on the quantised models. The quantisation brings the size of models down to between 588 bytes and 68 bytes. For complex textures involving lots of fine detail without much regularity, the model degrades gracefully with size, in most cases continuing to capture macroscopic detail (colour distributions and larger features). Simpler and more regular patterns, such as the zebra stripes, are captured equally well by both the small and large models. 

\begin{table}
\centering
\begin{tabular}{c@{\hskip 0.1cm}c@{\hskip 0.1cm}c@{\hskip 0.1cm}c@{\hskip 0.1cm}}
    TNCA & $\mu$NCA & \textbf{Ours} & \\ 
    VGG-loss & VGG-loss & OTT-loss & \\ 
    5856 params & 588 params & 588 params & Target \\ 
    \phototwo{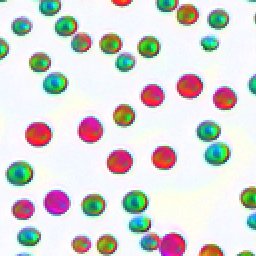} & \phototwo{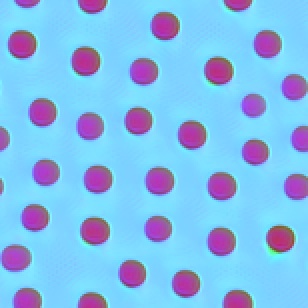} & \phototwo{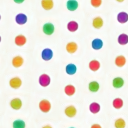} & \includegraphics[height=2cm]{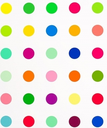} \\
    \phototwo{viking_VGG} & \phototwo{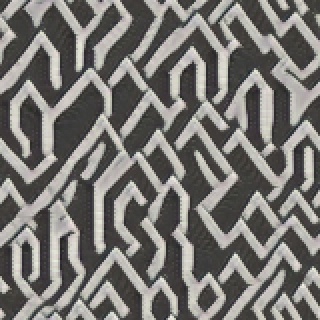} &  \phototwo{viking_OTT} & \includegraphics[height=2cm]{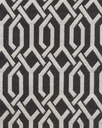} \\
    \phototwo{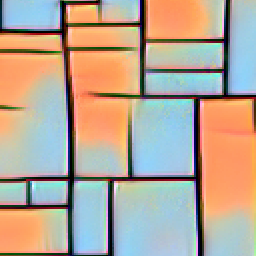} & \phototwo{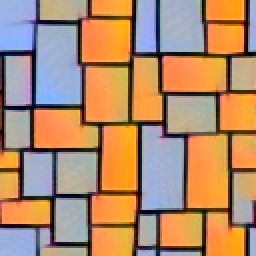} &   \phototwo{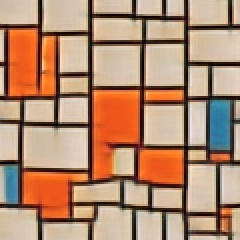} & \includegraphics[height=2cm]{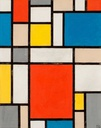}\\
\end{tabular}
\captionof{figure}{\label{tab:ott_knobs}Comparison with original TNCA loss and architecture (VGG-loss and 5856 parameters) and with $\mu$NCA trained using VGG-loss (588 parameters). A persistent problem with matching image statistics in a pre-trained convolutional network is drift in the colour channels and inaccurate colour reproduction, as seen here. In TNCA, it appears this problem is somewhat alleviated by the relatively higher parameter count - applying VGG-loss to the 588 parameter $\mu$NCA degrades the quality of the produced patterns in most cases. OTT-loss is unbiased in colour space, as each primary colour channel is simply a dimension in the patch feature space and treated equally under optimal transport divergence. We note that even spatially complex patterns, such as the weave, are still reproduced well by the $\mu$NCA, despite having far fewer parameters than TNCA.}
\end{table}

\begin{table}
\centering
\begin{tabular}{cccc}
    &  & \textbf{Ours} & \\ 
    &  &  588 params & Target \\ 
    \multirow{2}{*}{\rotatebox[origin=c]{90}{\textbf{(a)}}} & \phototwo{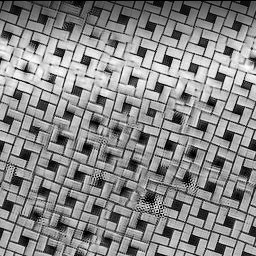} &  \phototwo{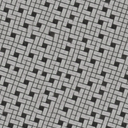} & \includegraphics[height=2cm]{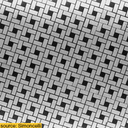} \\
     &\phototwo{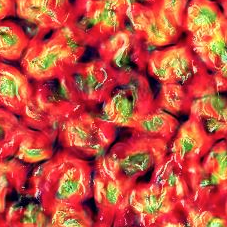} &  \phototwo{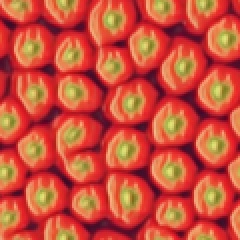} & \includegraphics[height=2cm]{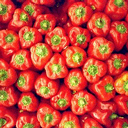} \\
    \midrule
    \rotatebox[origin=c]{90}{\hspace{18 mm}\textbf{(b)}}\vspace{-9 mm} &\phototwo{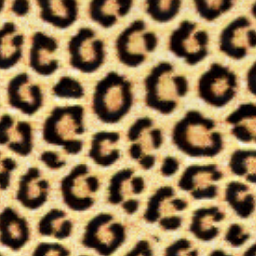} &  \phototwo{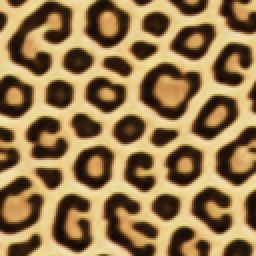} & \includegraphics[height=2cm]{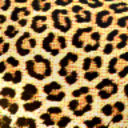} \\
    \midrule
    \multirow{2}{*}{\rotatebox[origin=c]{90}{\textbf{(c)}}} & \phototwo{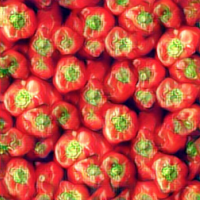} &  \phototwo{comparisons/peppers_OTT} & \includegraphics[height=2cm]{comparisons/peppers_orig} \\
    &\phototwo{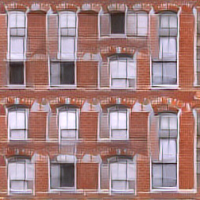} &  \phototwo{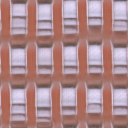} & \includegraphics[height=2cm]{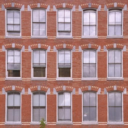} \\
    \midrule
    \rotatebox[origin=c]{90}{\hspace{18 mm}\textbf{(d)}}\vspace{-9 mm} & \phototwo{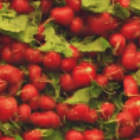} &  \phototwo{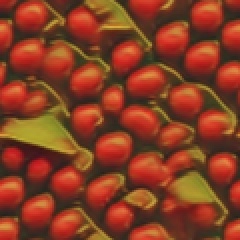} & \includegraphics[height=2cm]{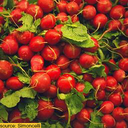} \\
    & & & 
\end{tabular}
\captionof{figure}{\label{tab:ott_knobs}Comparison of $\mu$NCA with different well-known texture synthesis approaches. In order, \textbf{a)} \emph{(710 params)} Portilla and Simoncelli's \cite{Portilla2000-ok} approach based on matching hand-defined pixel statistics, \textbf{b)} \emph{(parameter-free)} Hugues Hoppe's \cite{Lefebvre2006-kh} work on constructing and utilizing an "appearance space" to aid in the selection of patches from the target image, \textbf{c)} \emph{(approx. 65k params)} Ulyanov et al's work \cite{Ulyanov2017-af} on using pre-trained convolutional networks as feature-extractors to differentiably train a generator network, \textbf{d)} \emph{(approx. 65k params)} Houdard et al's work \cite{Houdard2020-qd} on using optimal transport to match feature statistics in images, in order to backpropagate into a generator network (as in (c)). We note they describe their generator network, TexNet, as "small", at 65k parameters.}
\end{table}



\subsubsection{Quantitative evaluation}

Quantitative comparison of a synthesized texture with the intended target is an ongoing area of research \cite{Lin2006-vc}, in no small part due to the open-ended definition of the texture synthesis task itself. Approaches from the family of Fréchet inception distances ("FID score"), commonly used in other image-generation settings, are not applicable, in part because the objective of $\mu$NCA is not to produce a competitive natural image generator, and also because most baselines for texture-synthesis, such as TNCA, utilize Gatys et. al approach of matching feature statistics during training. Thus, FID would make the optimization objective and evaluation criteria one and the same (albeit with a different convolutional architecture). Other works \cite{Lefebvre2006-kh,Houdard2020-av,Portilla2000-ok} in the field likewise present scant quantitative evaluation, and instead focus on qualitative samples and analysis.

\begin{table}
  \centering
  \begin{tabular}{lrr}
    \toprule
    $(N_{lap}, N_x, N_y)$ & C & $N_{params}$ \\
    \midrule
    (2, 1, 1) & 4 & 68 \\
    (2, 2, 2) & 6 & 150 \\
    (4, 2, 2) & 8 & 264 \\
    (4, 4, 4) & 12 & 588 \\
    \bottomrule
  \end{tabular}
  \caption{$\mu$NCA filter configurations used in the experiments.}
  \label{tab:config}
\end{table}

\subsection{Ablations}

We qualitatively study the ablation of two elements of our proposed training procedure - gradient normalisation and overflow loss. Samples of two patterns under these ablations can be seen in \ref{fig:ablations}. In addition to the perceptual degradation in quality, the training process exhibited instabilities under both ablations, with frequent explosions in state space or in gradient space. Such instabilities are exacerbated by use of the  training sample pool, where corruption of the states involved in a one training iteration may be re-sampled at later training iterations.

\begin{table}
\centering
\begin{tabular}{ccc} 
    \pbox{20cm}{\small{\textbf{Ours}} \\ \\} & \pbox{20cm}{\small{w/o} \\ \textbf{\small{gradientNorm}} \\} & \pbox{20cm}{\small{w/o} \\ \small{\textbf{gradientNorm}} \\ \textbf{\small{overflowLoss}}} \vspace{3mm} \\
    \phototwo{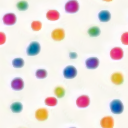} & \phototwo{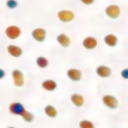} & \phototwo{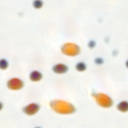}  \\
    \phototwo{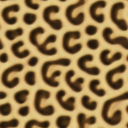} & \phototwo{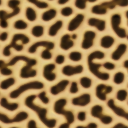} & \phototwo{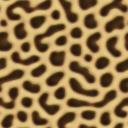} \\
\end{tabular}
\captionof{figure}{\label{fig:ablations} Qualitative ablation results for training augmentations introduced. In addition to degradation in the quality of the converged pattern, lack of these augmentations often leads to exploding states, or exploding gradients.}
\end{table}

\subsection{Optimal Transport Loss Control}

The use of patches and an Optimal Transport based loss affords more fine grained control over the texture similarity metric than traditional feature-statistic matching based methods. The target texture is encoded as a distribution of image patches; tuning the size and nature of these patches provides intuitive "control knobs" for the training process. For instance, the effective size of the largest patch at the highest level of the pyramid ($K*2^{N_{levels}-1}$) approximately determines the size of the largest features in the target image that will synthesized in the source image under generation. Note that this is only an approximate limit; the overlap between patches can still result in larger visual features being correctly synthesized, but more often than not patches will be recombined in a locally-consistent, globally-inconsistent manner. See figure \ref{tab:ott_knobs} to see the effects of different patch sizes.


\begin{table}
\centering
\begin{tabular}{cc}
    \multicolumn{2}{c}{\includegraphics[width=1.75cm]{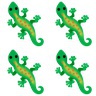}}\\
    \multicolumn{2}{c}{\textbf{target}}\\
    \midrule
    \photo{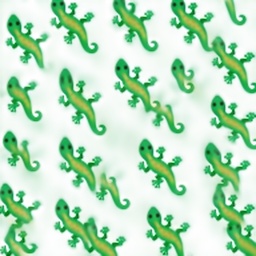} &  \photo{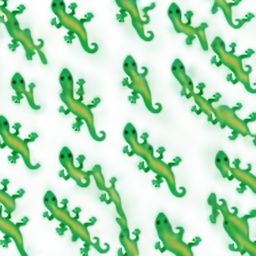} \\
    $K=7, N_{levels}=5$ & $K=5, N_{levels}=5$ \\ 
    \photo{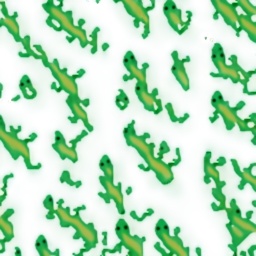} &  \photo{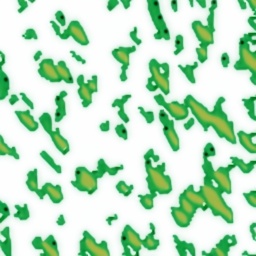} \\
    $K=3, N_{levels}=5$ & $K=3, N_{levels}=3$ \\
\end{tabular}
\captionof{figure}{\label{tab:ott_knobs}\textbf{Pixel-optimisation} using OTT-Loss using Adam with $1e-3$ learning rate for $200$ iterations. \emph{Note that $\mu$NCA are unable to learn this target image; we intend this figure to demonstrate the "control knobs" made available by OTT-Loss}. Larger patches and more levels result in consistent macroscopic detail, such as the structure of the body, while smaller patches and fewer levels produce an image with consistent microscopic detail but lacking all larger-scale coherence, resulting in an abundance of lizard eyes and other immediately local features, but no discernible structure to their composition.}
\end{table}
    
\subsection{$\mu$NCA Dynamics}

The inherent nature of $\mu$NCA as iterative models prevents static visualisations of their states from fully capturing their behaviour. We suggest readers interact with sample $\mu$NCA models, implemented as shader programs, available here: \href{https://www.shadertoy.com/view/slGGzD}{NCA-68}, \href{https://www.shadertoy.com/view/styGzD}{NCA-588}. Additionally, we provide code to train $\mu$NCA in the supplementary materials.

\section{Discussion}

\subsection{Procedural Generation}

We hypothesize that the extreme constraint on the number of parameters prevents overfitting or even learning any significant section of the bitmap verbatim (indeed, the target sample image bitmaps themselves are approximately $128 * 128 * 3 = 49152$ bytes, as compared to 68 bytes in our smallest model). Similarly, see figure \ref{tab:jpeg_comparison} to see a comparison to a traditional image encoding algorithm, JPEG, with the lowest quality setting at approximately 500 bytes versus 264 bytes in our smallest quantised model, albeit with some overhead for the JPEG file format. Other formats, such as JPEG-XL (JXL), allow for far more compact representations, and as a result there is a field of "JXL art" \cite{noauthor_undated-li} focusing on writing minimal valid JXL files on the order of tens of bytes. At first glance, these hand-engineered JXL files appear to match the expressivity of $\mu$NCA. However, the underlying model of $\mu$NCA could conceivable be implemented in at most hundreds of lines of standard C code \cite{Mordvintsev_undated-hy}, while the JPEG-XL reference library \cite{noauthor_undated-ty} is composed of hundreds of thousands of lines. 

We posit that the learned parameters represent a near-minimal procedural program that satisfies the desired behaviour (to stochastically generate a perceptually similar texture to a given sample image), and that the model must learn to exploit any regularities in the pattern to achieve this task. Hand-engineered procedural texture generators are often written in an extremely compact fashion, but still rely on extensive graphics libraries. We emphasize that most of the generative process, or the procedural program, for a learned texture is effectively entirely encoded in the parameters; the underlying architecture used for $\mu$NCA is not complex.

\begin{table}
\centering
\begin{tabular}{cc}
    \photo{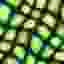} &  \photo{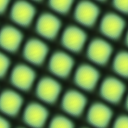} \\
    JPEG $Q=4\%$, 500 bytes & $\mu$NCA, 264 parameters \\ 
\end{tabular}
\captionof{figure}{\label{tab:jpeg_comparison}Comparison of a JPEG encoded image and the output of a $\mu$NCA rule with fewer than half the number of bytes. $\mu$NCA make use of regularity in patterns to learn procedural generation.}
\end{table}

\begin{table}
\centering
\begin{tabular}{c@{\hskip 0.1cm}c@{\hskip 0.1cm}c@{\hskip 0.1cm}c}
    \includegraphics[trim={0 7cm 7cm 0},clip,width=1.9cm]{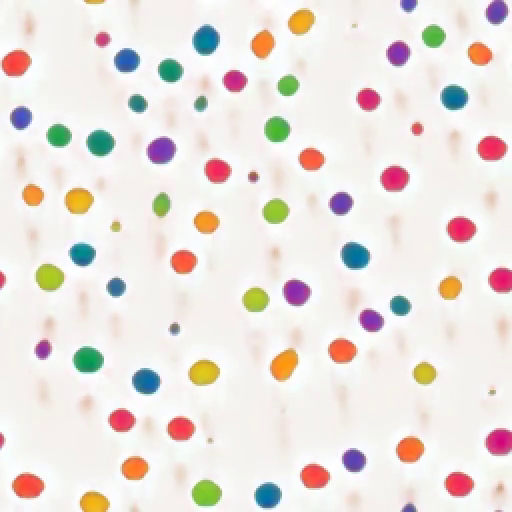} &
    \includegraphics[trim={0 7cm 7cm 0},clip,width=1.9cm]{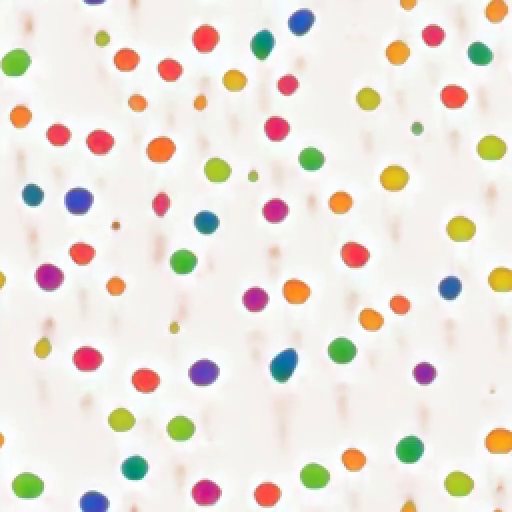} &
    \includegraphics[trim={0 7cm 7cm 0},clip,width=1.9cm]{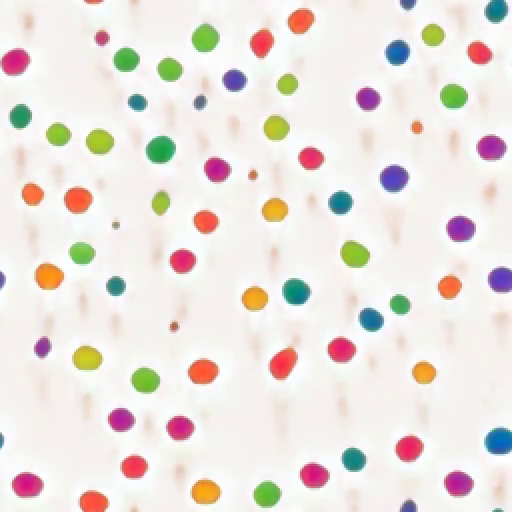} &
    \includegraphics[trim={0 7cm 7cm 0},clip,width=1.9cm]{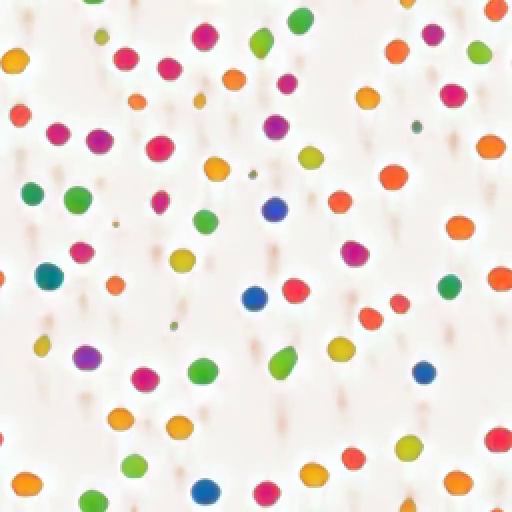} \\
    $t$ & $t+10$ & $t+20$ & $t+30$
\end{tabular}
\captionof{figure}{\label{tab:oscillating}Four samples, each $10$ steps apart, after pattern convergence, from a $\mu$NCA trained on a polka dot pattern. Occasionally, the $\mu$NCA converges to the remarkable procedural solution of synthesizing a number of dots, then having each dot oscillate through the target colours, while ensuring they are not in sync in order to produce the desired target distribution of colours among the dots at any given time-step. Closer inspection reveals the dots oscillate across colours in the same order.}
\end{table}

 We find further evidence for this hypothesis when qualitatively observing the model's behaviour during iteration. When the model is trained on a "polka dot" pattern \ref{tab:oscillating}, which features regularly spaced dots of different colours against a white background, we observe that on occasion the $\mu$NCA with 8 channels (280 parameters) will converge to a solution where the coloured dots oscillate in colour, as opposed to simply generating an image with several multi-coloured dots spatially distributed across the image in different, fixed, colours. We believe this behaviour is indicative that the solution space of $\mu$NCA is the space of possible procedural programs, in contrast to some parameterisation of different statistics of the image. Designing a program where the output is a polka dot pattern in different colours, with the constraint that the program's output must be a valid "polka dot" pattern at any given step after some period of convergence, certainly admits the non-intuitive solution of producing bubbles of oscillating colour.

\subsection{Limitations}


The method outlined here, like the originally proposed NCA, learns a complete generative model per target texture. This is comparable to single-image CPPNs and other work that uses neural networks to encode single data points \cite{Stanley2007-hc} \cite{Ulyanov2017-af}. It is conceivable that knowledge could be shared between different models operating on the same space of natural images, and thus a current limitation which may need to be explored further is the sharing of parameters between models. However, the focus of this work on ultra-compact models narrowed the scope of experiments to those involving single images per model.

The texture samples produced by $\mu$NCA and the underlying work TNCA do not produce images indistinguishable from images from the natural world. The models perform significantly better on regular patterns (especially artifically produces ones, such as grids or other repeated features). The distinguishing features of $\mu$NCA are their extremely compact nature and robustness inherent to their architecture (embarrasingly parallel and highly fault-tolerant).

{\small
\bibliographystyle{ieee_fullname}
\bibliography{egbib}
}

\end{document}